\documentclass[runningheads]{llncs}
\usepackage[T1]{fontenc}
\usepackage{graphicx}
\usepackage{algorithm}
\usepackage{algpseudocode}
\usepackage{amsmath}
\usepackage{float}
\usepackage{soul}
\usepackage{xcolor}
\usepackage{orcidlink}

\sethlcolor{yellow} 

\begin{document}
\title{EfficientNet in Digital Twin-based Cardiac Arrest Prediction and Analysis}
%
%\titlerunning{Abbreviated paper title}
% If the paper title is too long for the running head, you can set
% an abbreviated paper title here
%
\author{Qasim Zia\inst{1}\orcidlink{0009-0004-2028-5960} \and
Avais Jan \inst{1}\orcidlink{0009-0006-0199-0984} \and
Zafar Iqbal \inst{1}\orcidlink{0009-0005-6139-0019} \and
Muhammad Mumtaz Ali \inst{2}\orcidlink{0000-0002-4709-1259} \and
Mukarram Ali \inst{3}\orcidlink{0009-0007-8470-1013} \and
Murray Patterson\inst{1}\orcidlink{0000-0002-4329-0234}}
\authorrunning{Q. Zia et al.}
% First names are abbreviated in the running head.
% If there are more than two authors, 'et al.' is used.
%
\institute{Georgia State University, Atlanta, GA, 30303 USA \and
Zhengzhou University, Zhongyuan District, Zhengzhou, Henan, China, 450001\and
Dublin Business School, 13/14 Aungier St, Dublin, D02 WC04, Ireland 
}
\maketitle              % typeset the header of the contribution
\begin{abstract}
Cardiac arrest is one of the biggest global health problems, and early identification and management are key to enhancing the patient's prognosis. In this paper, we propose a novel framework that combines an EfficientNet-based deep learning model with a digital twin system to improve the early detection and analysis of cardiac arrest. We use compound scaling and EfficientNet to learn the features of cardiovascular images.  In parallel, the digital twin creates a realistic and individualized cardiovascular system model of the patient based on data received from the Internet of Things (IoT) devices attached to the patient, which can help in the constant assessment of the patient and the impact of possible treatment plans. As shown by our experiments, the proposed system is highly accurate in its prediction abilities and, at the same time, efficient. Combining highly advanced techniques such as deep learning and digital twin (DT) technology presents the possibility of using an active and individual approach to predicting cardiac disease.

\end{abstract}
\section{Introduction}
Cardiovascular disease (CVD) is the leading cause of death worldwide, and arrest is one of the most deadly manifestations \cite{celermajer2012cardiovascular}. Arrest is a significant and serious complication of ischemic heart disease, and early and accurate prediction is crucial for effective management and potential improvement in patient outcomes \cite{schultz2015strategies}. The application of artificial Intelligence (AI) and Digital Twin (DT) in healthcare has opened new opportunities for patient monitoring and predictive analysis in real time \cite{jameil2024digital}.  DT emulates a patient's cardiovascular system based on real-time analysis of patient data, helping to provide an accurate risk assessment and early identification of arrest events \cite{jemaa2023digital}. However, traditional machine learning and deep learning models are limited by their computational complexity and inefficiency in real-time analysis, which hinders their practical application in critical healthcare settings \cite{bian2022machine}.

Due to its optimal architecture and the ability to scale, EfficientNet, a state-of-the-art deep learning model, has shown excellent performance in medical image analysis \cite{ahmed2024optimizing}. CNNs, conventional convolutional neural networks, lack compound scaling, a systematic balance of depth, width, and resolution that EfficientNet employs to improve predictive accuracy while maintaining computational efficiency \cite{lin2023efficient}. This is ideal for cardiac arrest prediction and analysis in the DT environment, where real-time performance and accuracy are crucial \cite{shah2021smart}.

Although EfficientNet has not been applied to digital twins to predict cardiac arrest, there is a gap in the literature regarding their integrated application. Using EfficientNet with high accuracy and efficiency in a digital twin framework can provide a new way of real-time risk assessment and simulation of cardiac events. This integration could improve predictive performance by combining robust image analysis with dynamic, patient-specific modeling, particularly with emerging federated learning and privacy-preserving data-sharing techniques.

The primary objective of this study is to create an EfficientNet framework to predict and analyze cardiac arrest in the DT environment as mentioned below:

\begin{itemize}
\item To develop an EfficientNet-based deep learning model for the prediction of cardiac arrest from real-time patient data within a Digital Twin system.
\item Apply DT Technology to build a dynamic and individual virtual model of the cardiovascular system of a patient to improve predictive accuracy.
\item Assess the effectiveness of the proposed framework by comparing the performance of the proposed framework with conventional deep learning models in accuracy, precision, recall, and F1 score.
\item Improve computational efficiency and model interpretability for real-time clinical decision-making. Process data securely and reliably for cardiac health monitoring through private smart sensing through the DT framework.
\end{itemize}

This paper applies EfficientNet in a DT framework to predict cardiac arrest. Federated Transfer Learning, privacy-preserving analysis, and real-time data processing are used to increase diagnostic accuracy with reduced computational costs. The study indicates that EfficientNet could revolutionize predictive healthcare and improve patient outcomes through an early and precise cardiac risk assessment. The main contributions of this work are:
\begin{itemize}
    \item Propose DT-based EfficientNet framework for Cardiac Arrest Prediction.
    \item Optimization model to improve cardiac arrest prediction without wasting computational resources and time.
    \item Comparisons with other CNN models to highlight the accuracy and performance benefits of EfficientNet.  
\end{itemize}

This paper is organized as follows. Section \ref{sec:related_work} explains related work. The architecture of a prediction of cardiac arrest based on EfficientNet is suggested in Section \ref{sec:eff_based_dtcardiac}. The experiments are then carried out comprehensively and assessed in Section \ref{sec: eval} to determine the effectiveness and efficiency of our suggested architecture. We finally arrive at our conclusions in Section \ref{sec:concl}.

\section{Related Works}\label{sec:related_work}
Cardiac arrest is one of the most important global health problems and survival rates depend on early prediction and management of the disease. Conventional machine learning and deep learning models have been extensively applied for cardiovascular disease detection; however, their integration with DT technology and real-time assessment is novel. The Digital Twin framework offers real-time feedback, emulation, and prognostic evaluation of a patient's cardiac condition from patient to patient, thus assisting the clinician in decision-making. Several studies explored deep learning applications in cardiac health screening. 
\subsection{Deep Learning for Cardiac Arrest Prediction} 
Recent studies have found that deep learning can significantly improve the prediction of cardiac events. Handcrafted features-based classical machine learning models such as logistic regression or support vector machines have been replaced by deep neural networks in recent times. Elola et al. \cite{elola2019deep} used convolutional neural networks (CNN) and recurrent neural networks (RNN) to successfully extract spatial and temporal features from multimodal data such as electrocardiograms ECG and other physiological signals to improve the early detection and prognosis of cardiac arrest using them. Hannun et al. (2019) \cite{hannun2019cardiologist} found that Deep Neural Networks (DNNs) were able to identify arrhythmias with the accuracy of experts which opens up the possibility of applying AI-driven approaches to cardiac event prediction. Similarly, Rajpurkar et al. (2017) \cite{rajpurkar2017cardiologist} proposed the application of convolutional neural networks (CNNs) for the identification of abnormalities in ECG signals, thus supporting the notion of AI-based screening

\subsection{EfficientNet in Medical Image Analysis} 
EfficientNet, a family of convolutional neural networks, was designed to achieve high accuracy with minimal computational cost and can be applied effectively to medical image analysis problems.
EfficientNet was developed by Tan and Le \cite{tan2019efficientnet} and has revolutionized CNN architecture design using a compound scaling method that adjusts the depth, width, and resolution of the network uniformly. This innovative technique allows EfficientNet to achieve top-tier accuracy using fewer parameters and requiring fewer computational resources than previous models such as ResNet and DenseNet. Its success has been confirmed in medical imaging applications, such as tumor identification \cite{tripathy2023automation} and organ segmentation \cite{chekroun2023deep}, making it an attractive option to achieve performance and efficiency. Recent research, including that of Kaba et al. (2023) \cite{kaba2023application}, has also investigated the use of EfficientNet in cardiovascular imaging and has established that it provides better classification results than other conventional models. However, the integration of EfficientNet for the digital twin model of cardiac arrest remains obscure. 

\subsection{Digital Twin Technology in Healthcare} 
The application of digital twin technology implies the development of a virtual model of a system, specifically for continuous monitoring, simulation, and predictive analysis of a given physical system; in this case, it is the cardiovascular system of a patient \cite{coorey2022health}. Zia et al. \cite{zia2024priority} use a priority algorithm in a network based on Digital Twins (DT) to enhance communication. Awasthi et al.  \cite{awasthi2025exploring} discuss that Digital twins are currently employed in healthcare to model patient-specific dynamics from real-time data collected by wearable sensors, imaging modalities, and electronic health records. It enables personal treatment planning and real-time decision support in treatment, thus improving diagnostic accuracy and potential for early intervention in critical conditions such as cardiac arrest.\cite{martinez2019cardio} In this article, Martinez-Velazquez et al. (2019) explained the DT paradigm as a virtual copy of a patient's heart that can be constantly monitored and evaluated. When used in conjunction with rational use of features, this approach can enhance current methods of detection and management of the condition. Based on the studies reviewed above, the present research develops on previous approaches and presents a new concept of combining EfficientNet and Digital Twin technology. This integration is expected to improve the precision and speed of the cardiac arrest prediction model, thus solving a significant missing link in real-time healthcare surveillance.

\section{The Proposed EfficientNet in Digital Twin-based Cardiac Arrest Prediction and Analysis}\label{sec:eff_based_dtcardiac}
This section presents an end-to-end framework for integrating EfficientNet in a DT system for real-time cardiac arrest prediction and analysis. The methodology combines data-driven deep learning with patient-specific simulation to enable proactive clinical decision-making.

\subsection{Dataset}
\begin{figure*}[!ht]
    \centering
    \includegraphics[width=\textwidth]{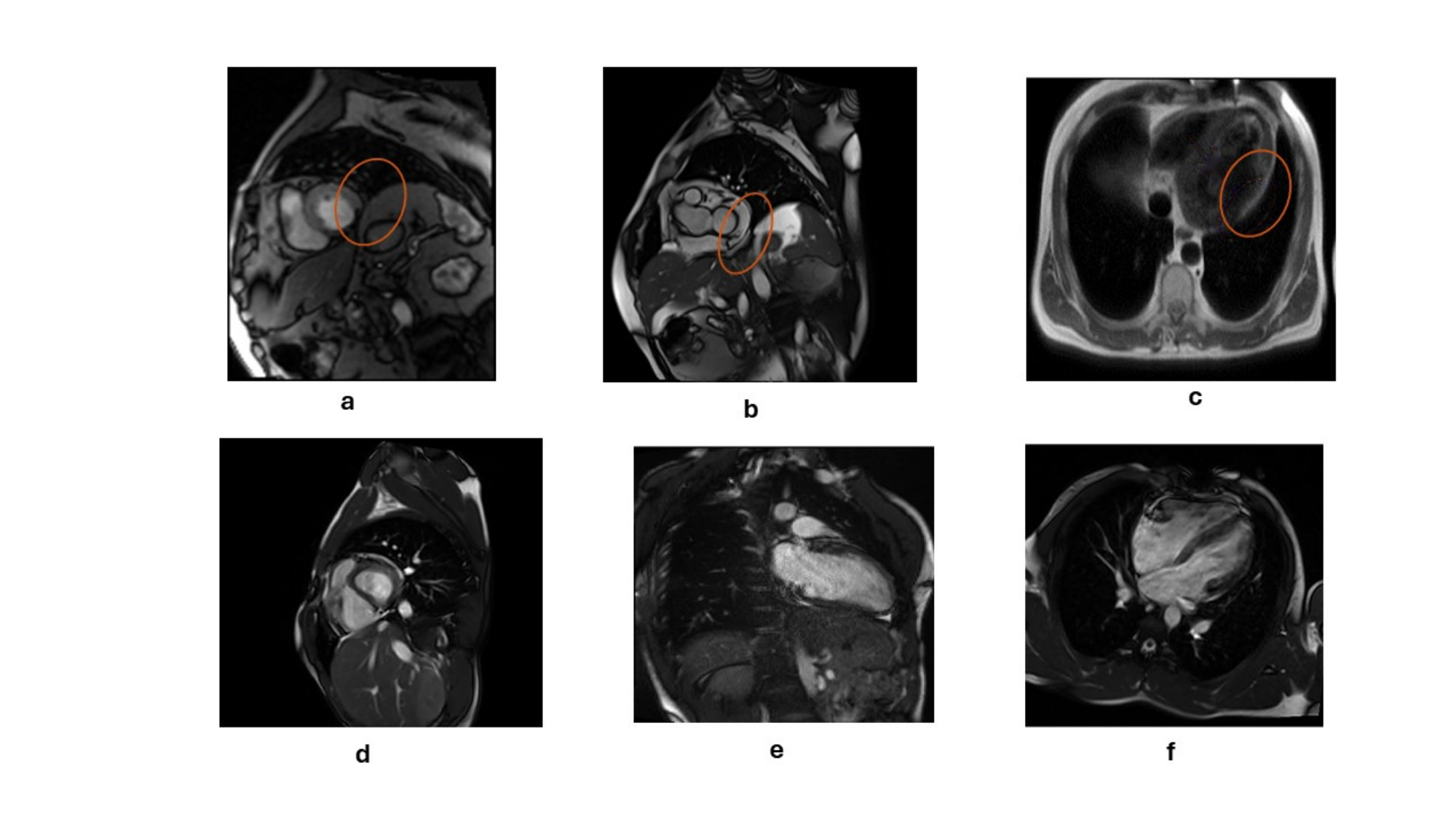}
    \caption{Sample Cardiac MRI images from dataset \cite{sharifrazi2022cardiac}}
    \label{fig_1}
\end{figure*}

We used the Cardiac MRI images\footnote{\scriptsize(\url{https://www.kaggle.com/datasets/danialsharifrazi/cad-cardiac-mri-dataset/data})} \cite{sharifrazi2022cardiac} dataset for our simulations. This is one of the most comprehensive Cardiac Coronary Artery Disease(CAD) image datasets. This dataset contains CAD disease images, which often lead to cardiac arrest. Figure \ref{fig_1} shows the images with CAD disease and normal images. (a-c) cardiac magnetic resonance images in Figure \ref{fig_1}  are from patients with coronary artery disease and (d-f) are healthy person images. The dataset has more than 60,000 images. The images in the dataset have different image quality. It is harder to identify the CAD in some images. So, data preprocessing is necessary. Training the model on heterogeneous image qualities leads to learning invariant features robust to noise and artifacts. This enhances the model's ability to identify subtle signs of Coronary Artery Disease (CAD) even when the imaging is not the best quality. Testing the model on a dataset with different quality is a more challenging task.  According to the studies by Litjens et al. (2017)  \cite{litjens2017survey} and Esteva et al. (2019) \cite{esteva2019guide} models that are trained on a large variety of image qualities are more likely to perform well. So such a trained model when tested on digital twin images which are often generated under ideal conditions with minimal noise and artifacts is more likely to perform well. Hence, this dataset provides a better platform to evaluate models than digital twin images, which are simple to detect and predict any disease.
\begin{figure*}[!ht]
    \centering
    \includegraphics[width=\textwidth]{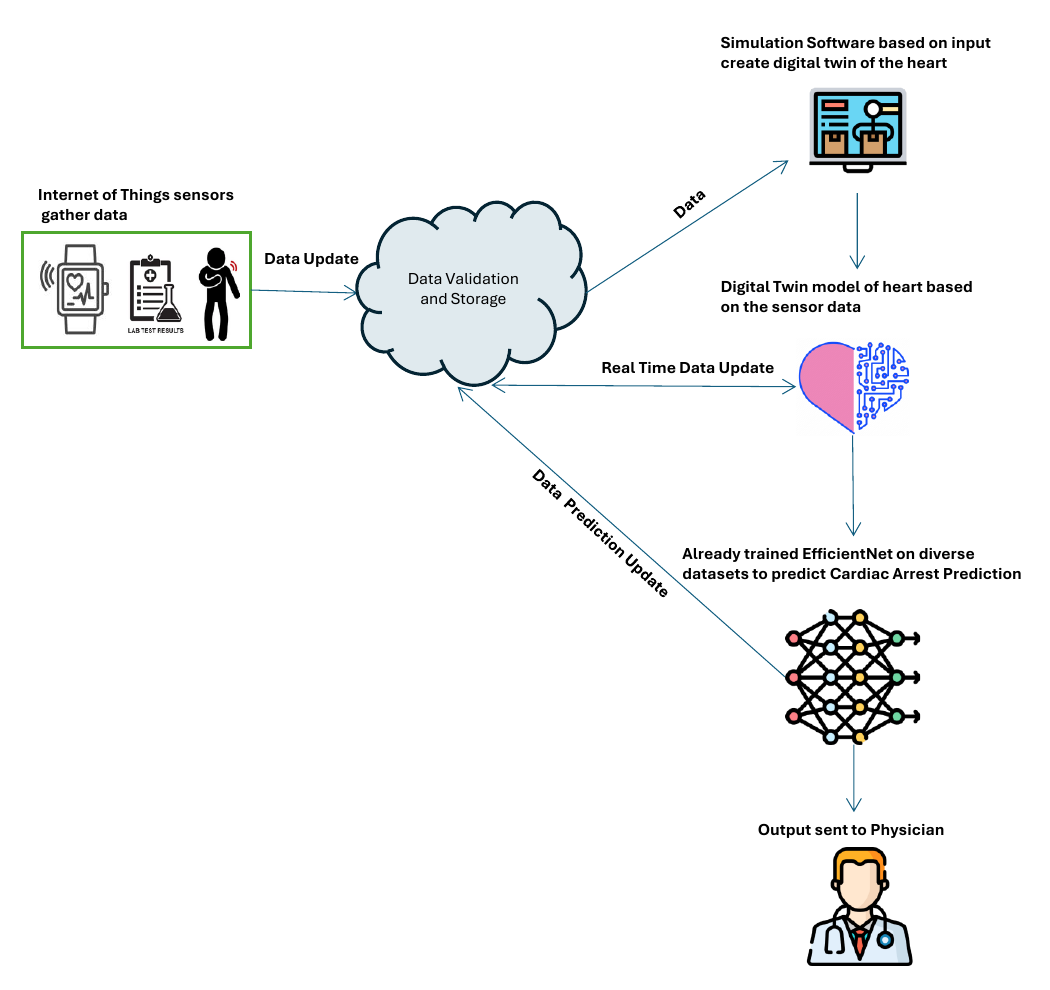}
    \caption{High-level architecture of using EfficientNet model for digital twin-based Cardiac Arrest Prediction and Analysis}
    \label{fig_2}
\end{figure*}

\subsection{Architecture of the Proposed EfficientNet Model for Cardiac Arrest}
This section provides a high-level architecture for using the EfficientNet model in DT-Cardiac Arrest Prediction and analysis, as seen in Figure \ref{fig_2}. The proposed framework consists of two main components:

 A digital twin is a virtual replica of the patient's cardiovascular system, and an EfficientNet-based prediction model takes clinical data as input to predict cardiac arrest risk. Data from wearable sensors, clinical imaging, and electronic health records (EHR) are continuously fed into the digital twin, and the EfficientNet model learns the relevant features to make real-time predictions.
 
The following is the proposed model workflow:
\begin{itemize}
    \item \textbf{Step 1:  Patient data (e.g., IoT wearable devices sensor data) is collected and uploaded to the cloud.}
    \item \textbf{Step 2: After Data Preprocessing}, the cloud runs the digital twin simulation software, which creates a 3D model of the cardiovascular system and simulates its behavior
    \item \textbf{Step 3: The simulation results are passed through EfficientNet-based AI models to predict cardiac arrest risk or other cardiovascular conditions}
    \item \textbf{Step 4: The predictions and insights are sent back to the healthcare providers or hospital staff for further analysis and decision-making.}
\end{itemize}

 \subsection{ IoT Wearable Device Data Collection and Transmission for Digital Twin-based Cardiac Arrest Prediction}
The IoT Wearable Device Data Collection and Transmission Algorithm is designed to predict cardiac arrest in real time using digital twins. First, it gathers real-time sensor data \( S = \{s_1, s_2, \ldots, s_n\} \) from wearable devices like heart rate monitors, blood pressure sensors, and other IoT-based health trackers. This data is combined with other environmental factors like location and activity level. 

The data is then sent to the cloud server over an existing communication network (4G, 5G, WiFi, etc.). The transmission process is divided into nested loops to ensure that data from different wearable devices and their neighbors are processed sequentially. For each wearable device \( W \), data collection and transmission take place over a time window \( t \) for all the available neighboring devices \( N \). The sensor data \( S_W(t) \) that is raw transmitted to the cloud for further analysis. After processing each neighboring device, it guarantees that the next device is updated by \( N \leftarrow N + 1 \). When all the neighboring devices are processed, the next wearable device is considered by \( W \leftarrow W + 1 \).

In this manner, systematic and real-time monitoring and data transmission is enabled to construct digital twin-based cardiac stroke prediction. The data allows cloud-based analysis and thus predicts early cardiac arrest detection and personalized healthcare interventions. The model uses these inputs to update the digital twin and continuously monitor the patient's heart health. Therefore, the algorithm designed for wearable devices and their data collection and transmission is essential for effectively realizing the concept of a digital twin in predicting and managing cardiac arrests.
\begin{algorithm}
\caption{ IoT Wearable Device Data Collection and Transmission for Digital Twin-based Cardiac Arrest Prediction}
%\begin{scriptsize} % Adjust font size to scriptsize
\begin{algorithmic}[1]
\State \textbf{Input:} Real-time sensor data \( S = \{s_1, s_2, \ldots, s_n\} \) from wearable devices
\State \textbf{Output:} Preprocessed data sent to the cloud server for further analysis
\State \textbf{Step 1: Data Upload Initialization}
\State Each wearable device \( W \) collects real-time sensor data \( S_W \) (e.g., wearable device, laboratory results, blood pressure )
\State The wearable device uses the existing communication network (e.g., 4G, 5G, WiFi) for data transfer to the cloud server
\State Data includes individual readings and environmental context (location, activity level)
\State Initialize wearable device \( W \), neighbor devices \( N \), and time \( t \)
\While{\( W \leq |W| \)}
    \While{\( N \leq N \)}
        \While{\( t \leq T \)}
         \State \textbf{Step 2: Transmit Data to Cloud Server}
         \State Send raw data \( S_W(t) \) from wearable device \( W \) to the cloud for analysis   
        \EndWhile 
    \State \textbf{Update for next iteration:}
    \State Update \( N \leftarrow N + 1 \) for the next neighboring device
    \EndWhile
\State Update \( W \leftarrow W + 1 \) for the next wearable device
\EndWhile

\end{algorithmic}
%\end{scriptsize}
\end{algorithm}

\subsection{EfficientNet in Digital Twin-based Cardiac Arrest Prediction and Analysis}
\subsubsection{ Cloud Server Side Cardiac Arrest Prediction Using EfficientNet}
The Cloud Server Side Cardiac Arrest Prediction Algorithm uses the EfficientNet model and DT technology to predict the risk of cardiac arrest using real-time sensor data from wearable devices.
\begin{itemize}
\item Data Preprocessing: This paper uses normalization of the raw sensor data \( S_W(t) \) obtained from wearable devices to use the data for analysis. The normalization process guarantees that the data is within a certain range of values, which will help produce more accurate predictions in the subsequent stages.
\item Digital Twin Creation: A Digital Twin (DT) is simulated and is created using the processed sensor data. This virtual model uses the incoming sensor data to indicate the real-time condition of the Cardiovascular system, such as blood flow and electrical activities.
\item Integration into Digital Twin Model: The sensor data is integrated into the  Cardiovascular system Digital Twin model. It enables the simulation of the  Cardiovascular system dynamic conditions, which is important for precise cardiac arrest risk prediction.
\item Cardiac Arrest Prediction using EfficientNet: The algorithm uses an already trained EfficientNet model for cardiac arrest risk prediction. It processes the cardiovascular system DT model and produces a cardiac arrest risk prediction \( \hat{DS_W}(t)\).
\item  Feed-forward Pass for Prediction: To determine the cardiac arrest risk for each patient, we perform a feed-forward pass through the EfficientNet model, using their respective Digital Twin models as input.
\item Cardiac Arrest Prediction Decision: Based on the processed data, a cardiac arrest prediction decision is suggested in real-time for the patient.
\item Output Sent to Physician: The cardiac arrest prediction results and the DT model are delivered to the physician for further review. It enables the physician to evaluate the cardiac arrest risk and make appropriate decisions.
\item Anomaly Detection: The predicted decision \( \hat{DS_W}(t)\) is then compared with the actual cardiac arrest occurrence outcome \( {DS_W}(t)\) to detect anomalies using a 3$\sigma$ rule, following the receipt of feedback. This compares it over time to improve the prediction model continuously.
\item Weighted Decision Making: The decision-making process is further enhanced by incorporating the output of the patient and the physician through a weighted average. The algorithm uses the weights to weigh the feedback from one source or another depending on its relevance. where \(\alpha\) is the weight for patient feedback, and \(w_{Wn}\) are the weights for doctor feedback.
\item Fine-Tuning the EfficientNet Model: The EfficientNet model is fine-tuned based on the feedback received for better accuracy in future predictions.
\item Cloud Server Update: The cloud server state is updated depending on the feedback and predictions to optimize the model for the next cardiac arrest risk prediction iteration.
\end{itemize}
The real-time cardiac arrest prediction algorithm identified above leverages wearable device data, digital twin simulations, and EfficientNet for accurate decision-making and offers significant insights for medical professionals.
\newcommand{\Input}[1]{\item[\textbf{Input}] #1}
\newcommand{\Output}[1]{\item[\textbf{Output}] #1}
\begin{algorithm}[H]
\caption{Cloud Server Side Cardiac Arrest Prediction Using EfficientNet}
\begin{scriptsize} % Adjust font size to scriptsize
\begin{algorithmic}[1]
\Input: Raw data from wearable device \( S_W(t) \);
\Output: Arrest prediction decision \( \hat{DS_W}(t) \) for the respective wearable devices \( W \) of the patient;
\State \textbf{Step 1: Pre-Processing and Managing Data at the DT Layer}
\[
S_W(t) \leftarrow \frac{S_W(t)}{\sigma}
\]
\State \textbf{Step 2: Simulation Software to Create a DT of the Cardiovascular system}
\[
DT_{\text{CVS}}(t) = \text{Simulate}(S_W(t))
\]
\State \textbf{Step 3: Create DT Model of the Cardiovascular System using Sensor Data}
\[
DT_{\text{CVS}}(t) \leftarrow \text{Integrate}(S_W(t))
\]
\State \textbf{Step 4: EfficientNet Model Processing}
\[ \hat{DS_W}(t) = \text{EFF-NET}(DT_{\text{CVS}}(t)) \].

\State \textbf{Step 5: Feed-forward Pass for Prediction}
\State \textbf{Step 6: Suggest Cardiac Arrest Decision}

\State \textbf{Step 7: Output Sent to Physician}

\[
\text{Send Output to Physician: } \hat{DS_W}(t), DT_{\text{CVS}}(t)
\]

\State \textbf{Step 8: Feedback for Anomaly Detection}
\[
\text{Anomaly Detection: } \hat{DS_W}(t) \text{ vs. } DS_W(t) \text{ based on 3$\sigma$ rule}
\]
\State \textbf{Step 9: Weighted Decision Making}
\[
\hat{DS_W}(t) = \alpha \hat{DS_W}(t) + (1 - \alpha) \sum_{n=1}^{N} w_{Wn} \hat{DS_n}(t)
\]

\State \textbf{Step 10: Fine-Tune the EfficientNet Model}
\State \textbf{Step 11: Update Cloud Server State}

\State \textbf{Update for next iteration}

\State \textbf{End of algorithm.}
\end{algorithmic}
\end{scriptsize}
\end{algorithm}

\section{Performance Evaluation}\label{sec: eval}

This section includes a description of the experimental setup, the metrics used for evaluation, the quantitative results, and the comparison based on EfficientNet for predicting cardiac arrest.

\subsection{Baseline Studies}
The CNNs VGG16, ResNet50, and EfficientNet will be compared. VGG16 is renowned for its simplicity and efficacy, while ResNet50 has a contemporary architecture and is renowned for its performance and efficiency. TensorFlow is used in this work to implement all models. TensorFlow is used for training all models. The primary goal of our research as mentioned in the Introduction Section is to detect diseases and predict cardiac arrest in real-time. It is not affected by wireless transmission effects (such as network latency or capacity limitations) once we get the images depicting the patient's cardiovascular system.
The impact of the CNN model on the decision-making process is challenging to identify if we consider the additional variables introduced by wireless transmission (latency, packet loss, bandwidth restrictions, etc.).

\subsection{Experimental and Performance Analysis }
Several essential performance criteria will be used to evaluate the effectiveness of the EfficientNet model for predicting cardiac arrest. 

Accuracy: The percentage of data our model accurately predicts its accuracy. It is the proportion of accurately predicted values to all expected values.

Recall: The positive cases can be accurately identified by the model.

Specificity: The percentage of accurately identified negatives among all negatives is known as specificity.

F1 score: It determines the model's precision and recall harmonic mean. It calculates the metric by considering both precision and recall.
This section will discuss the performance of the proposed EfficientNet model for CAD detection which can lead to cardiac arrest. We also assess the time taken to predict. We will compare the EfficientNet model with VGG16 and ResNet50. EfficientNet performance is compared using various performance metrics that are
very important in predicting cardiac arrest, which includes accuracy, recall, F1 score, and Specificity in Table \ref{tab:table1}

The EfficientNet model's performance was compared with other models using various performance metrics, including accuracy, precision, recall, F1 score, AUC, and training time. These metrics are essential for decision-making and faster processing. 

The results show that EfficientNet outperforms the other model and provides better prediction. 

\begin{table}[htbp]
\footnotesize
\caption{\textbf{Performance Metrics for Different CNN Models} \label{tab:table1}}
\centering
\resizebox{\textwidth}{!}{%
\begin{tabular}{|l|c|c|c|c|c|c|c|c|c|}
\hline
\textbf{CNN Models} & \textbf{Accuracy (\%)}&\textbf{Precision} & \textbf{Recall} & \textbf{F1-Score} & \textbf{Specificity} & \textbf{Auc}  & \textbf{Training Time} \\
\hline
EfficientNet &93.45 & 0.925 & 0.9482 & 0.9365 & 0.9203 &  0.9630  & 275.24s \\
\hline
ResNet50 &91.26& 0.9077 & 0.9192 & 0.9134& 0.906 & 0.9487 & 320.40s \\
\hline
VGG16 &89.15& 0.8903 & 0.8933 & 0.8918 & 0.8896 & 0.9339 & 435.24s \\
\hline
\end{tabular}}
\end{table}

% \begin{figure}[htbp]
%     \centering
%     \includegraphics[width=\textwidth]{Cardiac confusion.pdf}
%     \caption{ Comparison of Confusion Matrices for EfficientNet, ResNet50 and VGG16}
%     \label{fig_3}
% \end{figure}

\begin{figure}[htbp]
    \centering
    \includegraphics[width=0.85\textwidth]{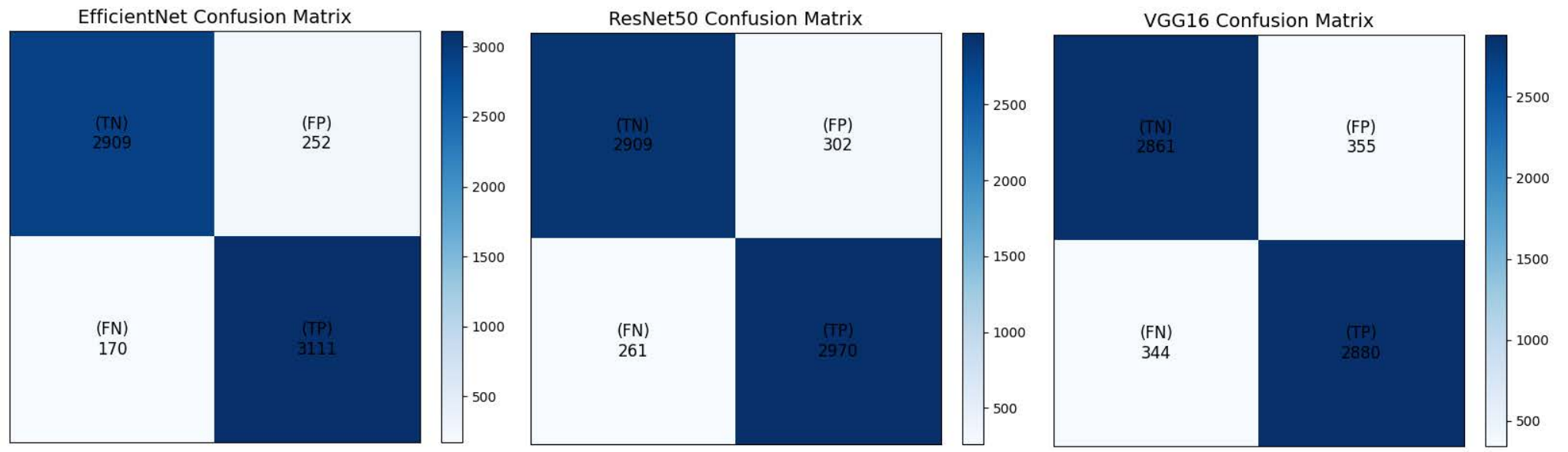}
    \caption{Comparison of Confusion Matrices for EfficientNet, ResNet50, and VGG16}
    \label{fig_3}
\end{figure}

The confusion matrix also evaluates EfficientNet, ResNet50, and VGG16 model performance. The details of the confusion matrix can be analyzed in Figure \ref{fig_3}. 
\section{Conclusion}\label{sec:concl}
In this paper, we present a new framework that uses an EfficientNet-based prediction model with a digital twin system to solve the critical problem of detecting early cardiac arrest. As such, we propose using EfficientNet for a dynamic patient-customized digital twin because of its ability to extract advanced features. Our approach creates a digital twin that offers a real-time assessment of the patient's cardiovascular system. The digital twin component, however, establishes a patient’s cardiovascular system simulation model through which the patient's current state can be depicted.

The results of our experiments show that EfficientNet can effectively predict the risk of cardiac arrest using cardiac image data. The proposed framework structure enables it to be easily extended to work in distributed healthcare settings. It has the potential to explore other innovations, like federated learning, for improved data privacy and collaboration across different institutions. 

Therefore, integrating EfficientNet and digital twin technology can significantly contribute to cardiac arrest prediction. Future work will include creating our digital twin model of the heart to confirm model generalization, optimizing model settings to improve the model's performance, and incorporating more sources of information to improve the system's accuracy. 
%
% ---- Bibliography ----
%
% BibTeX users should specify bibliography style 'splncs04'.
% References will then be sorted and formatted in the correct style.
%
% \bibliographystyle{splncs04}
% \bibliography{mybibliography}
%

\end{document}